\documentclass[10pt,twocolumn,letterpaper]{article}

\usepackage{wacv}
\usepackage{times}
\usepackage{epsfig}
\usepackage{graphicx}
\usepackage{amsmath}
\usepackage{amssymb}

\usepackage{subcaption}

\usepackage{pdfpages}

\usepackage{url}

\usepackage{array}
\newcolumntype{C}[1]{>{\centering\let\newline\\\arraybackslash\hspace{0pt}}m{#1}}

\wacvfinalcopy 

\begin{document}
\title{Eye-CU: Sleep Pose Classification for Healthcare using \\ Multimodal Multiview Data}

\author{Carlos Torres\textsuperscript{\dag} ~~~ Victor Fragoso\textsuperscript{\ddag} ~~~ Scott D. Hammond\textsuperscript{\dag} ~~~ Jeffrey C. Fried\textsuperscript{*} ~~~ B.S. Manjunath\textsuperscript{\dag} \\
\textsuperscript{\dag}Univ. of California Santa Barbara ~~ \textsuperscript{\ddag}West Virginia University ~~ \textsuperscript{*}Santa Barbara Cottage Hospital \\
{\tt\footnotesize{ \{carlostorres@ece, shammond@tmrl, manj@ece\}.ucsb.edu ~ victor.fragoso@mail.wvu.edu ~ jfried@sbch.org} }
}

\maketitle

\begin{abstract}
Manual analysis of body poses of bed-ridden patients requires staff to continuously track and record patient poses. Two limitations in the dissemination of pose-related therapies are scarce human resources and unreliable automated systems. This work addresses these issues by introducing a new method and a new system for robust automated classification of sleep poses in an Intensive Care Unit (ICU) environment. The new method, coupled-constrained Least-Squares (cc-LS), uses multimodal and multiview ($MM$) data and finds the set of modality trust values that minimizes the difference between expected and estimated labels. The new system, Eye-CU, is an affordable multi-sensor modular system for unobtrusive data collection and analysis in healthcare. Experimental results indicate that the performance of cc-LS matches the performance of existing methods in ideal scenarios. This method outperforms the latest techniques in challenging scenarios by 13\% for those with poor illumination and by 70\% for those with both poor illumination and occlusions. Results also show that a reduced Eye-CU configuration can classify poses without pressure information with only a slight drop in its performance.
\end{abstract}

\paragraph*{Keywords:}
\noindent Sleep poses, sleep analysis, patient positioning, coupled-constrained, Least-Squares optimization, multimodal, multiview, ICU monitoring, pose classification, healthcare, patient monitoring, modality contribution.
\section{Introduction}\label{sec:intro}
New innovative methods for non-disruptive monitoring and analysis of patient-on-bed body configurations, such as those observed in sleep-pose patterns, add objective metrics for evaluating and predicting health status. Clinical scenarios where body poses of patients correlate to medical conditions include sleep apnea -- where the obstructions of the airway are affected by supine positions \cite{sahlin2009sleep}. Mothers-to-be are recommended to lay on their sides to improve fetal blood flow \cite{morong2015sleep}. The findings of \cite{bihari2012factors, idzikowski2003sleep, weinhouse2006sleep} correlate sleep positions with various effects on patient health. In these studies, the findings highlight the importance of automated analysis of patient sleep poses in natural scenarios. They substantiate the need of this work and its potential benefits. The benefits include improving patient quality of life and quality of care through continuously monitoring patient poses, correlating poses to medical diagnosis, and optimizing treatments by manipulating poses. The proposed Eye-CU system and cc-LS fusion method tackles the classification of sleep poses in a natural ICU environment with conditions that range from bright and clear to dark and occluded. The system collects sleep-pose data using an array of RGB-D cameras and a pressure mat. The method extracts features from each modality, estimates unimodal pose labels, fuses unimodal decisions based on trust (priors) values, and infers a multimodal pose label. The trusts are estimated via cc-LS optimization, which minimizes the distance between the oracle and multimodal matrices. In this context, the term multimodal refers to the various Eye-CU sensor measurements.

\subsection{Related Work}\label{subsec:oldwork}
Computer vision methods using RGB data to detect body configurations of patients on beds are discussed in \cite{kuo2004artificial, liao2008video, penzel2000computer} but are limited to scenes with constant illumination and/or without occlusions. The deformable parts model approach, commonly used in RGB images presented in \cite{yang2013articulated} requires images with relatively uniform illumination and is limited to minor self-occlusions. The discriminative approach from \cite{shotton2013efficient} uses depth images and is robust to illumination changes. It requires clean depth segmentation and contrast and is susceptible to occlusions. A controlled method to classify human sleep poses using RGB images and a low-resolution pressure array is presented in \cite{huang2010multimodal}. It uses normalized geometric and load distribution features interdependently and requires a clear view of the patient.

The cc-LS work builds upon our previous work \cite{torres2015Multimodal}, where features from $R$, $D$, and $P$ sensors from a single view are combined to overcome challenging scene conditions. The trust method uses unimodal features to propose label candidates and infer a multimodal label. It improves unimodal decisions of Linear Discriminant Analysis (LDA) and Support Vector Classifier (SVC) via modality trust. Modality trust is defined as the mean classification accuracy of the unimodal pose classifiers (under the measured scene conditions). The trust system uses a high-resolution pressured mat and its performance relies heavily on a fixed camera over the patient's bed. A trust-adjustment method accounts for sensor failures; however, performance declines greatly without pressure data.

\subsection{Proposed Work}\label{subsec:propwork}
\begin{figure}[t]
    \begin{center}
     	\includegraphics[width=0.45\textwidth]{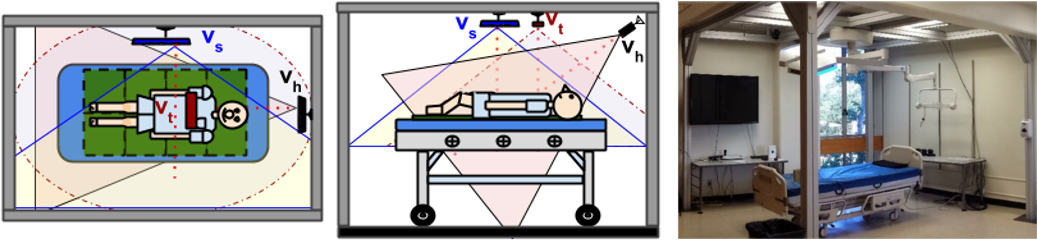}
    \end{center}
    \caption{\small Diagram of the Eye-CU physical setup showing the pressure mat (left) in green and the camera views(center): top ($v_t$) in red, side ($v_s$) in blue, and head ($v_h$) in black; and the mock-up ICU (right) where the system is tested.}
    \label{fig:MMsleepSetup}
\end{figure}

The work presented in this paper differs from \cite{torres2015Multimodal} by introducing a new probabilistic method to estimate trusts. We use cc-LS optimization (section \ref{sec:MMformula}) to estimate trusts, learn modality priors, and improve classification accuracy by up to 30\%. Instead of using a multimodal system with a single camera view and a pressure mat, the Eye-CU system uses multimodal and multiview ($MM$) data. Results suggest that combining reduced Eye-CU configurations with cc-LS robustly classifies sleep poses with incomplete views and without pressure information. Figure \ref{fig:MMsleepSetup} shows two perspective views of the system in the mock-up ICU room.

\begin{figure}[t]
    \begin{center}
		\includegraphics[width=.65\linewidth]{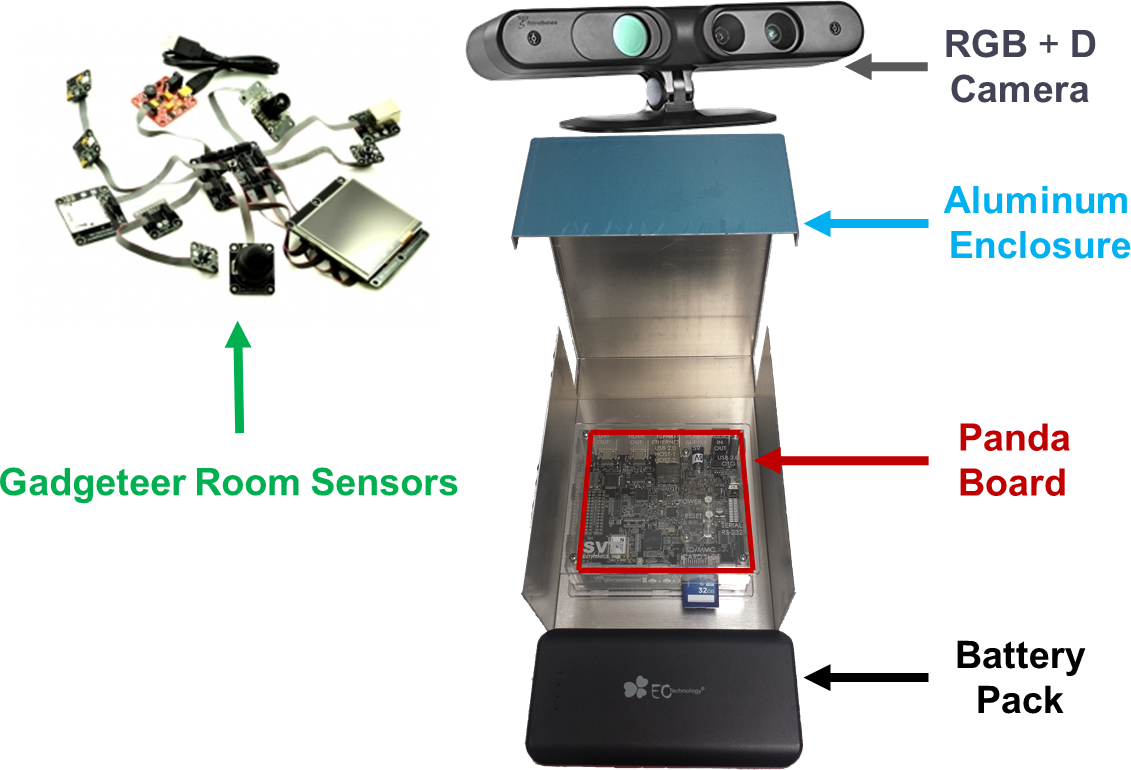}
    \end{center}
    \caption{\small Multimodal Eye-CU node with environmental sensors, RGB-D camera, aluminum enclosure, Panda Board, and battery pack. Four nodes are used to monitor a medical ICU room.}
    \label{fig:Node}
\end{figure}

\paragraph{Main Contributions of this work:}
(1) \textbf{cc-LS} a simple and elegant method to estimate modality trusts, which improves pose classification accuracy; (2) \textbf{Eye-CU} a complete modular $MM$ system that performs sleep pose classification with very high accuracy in healthcare. One node is shown in Figure \ref{fig:Node} and the system is currently deployed in a a medical ICU; and (3) a fully annotated $MM$ \textbf{dataset} of 66,000 sleep-pose images \footnote {will be available online at \textcolor{blue}{\url{http:vision.ece.ucsb}}}.

\section{Eye-CU System Description} \label{sec:System}
The various Eye-CU system configurations depend on the combination of modalities used: RGB ($R$), depth ($D$), and pressure ($P$), and available camera views: head ($h$), top ($t$), and side ($s$). The following configurations are explored:
\begin{itemize}
    \item Multimodal and Multiview ($MM$) uses $R, D, P$ data and the $h, s, t$ views. It is the most complex and has the best performance, but is difficult to deploy.
    \item Multimodal partial-Multiview ($MpM$) uses $RDP$ data and less than three views. $MpM$ with a top view is equivalent to the one used in competing methods.
    \item Partial-Multimodal and Multiview ($PMM$) uses $R$, $D$ or $RD$ data from three camera views ($hst$). Its performance depends on having all views available.
    \item Partial-Multimodal partial-Multiview ($PMpM$) is the simples configuration. It uses $RD$ data from two views ($hs, ht, st$) and sets the lower bound in performance.
\end{itemize}

\begin{figure}[t]
    \begin{center}
		\includegraphics[width=.65\linewidth]{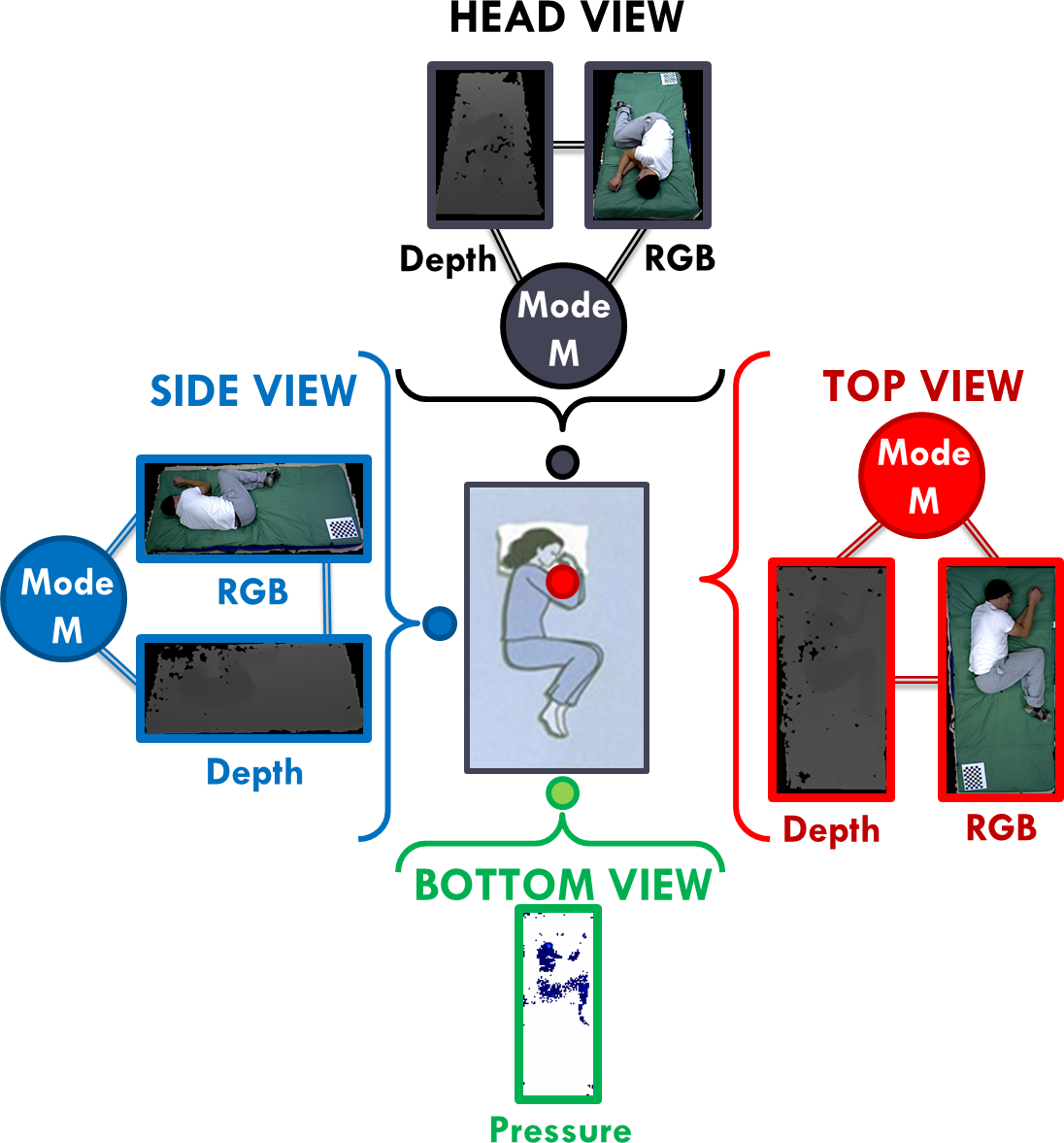}
    \end{center}
    \caption{\small Multimodal and multiview representation of the fetal left-oriented pose observed by three RGB-D cameras and one pressure-mat collected using the Eye-CU system.}
    \label{fig:MMposeRepresentation}
\end{figure}

\paragraph{Why Multimodal?}
Suitability tests (section \ref{sec:Experiments}) of existing methods and available modalities indicate that neither a single modality nor the concatenation of modalities can be used to classify poses in an natural ICU environment.
\paragraph{Why Multiview?}
The ICU is a dynamic environment, where equipment is moved around continuously and can block sensors and view of the patients. A multiview system improves classification performance, increases the chances of observing the patients, and enables monitoring using simple and affordable sensor. Cameras do not have contact with patients and avoid the risk of infections by touch.

\section{Data Collection}\label{sec:Data}
\begin{figure*}[t]
\begin{center}
	\includegraphics[width=.75\linewidth]{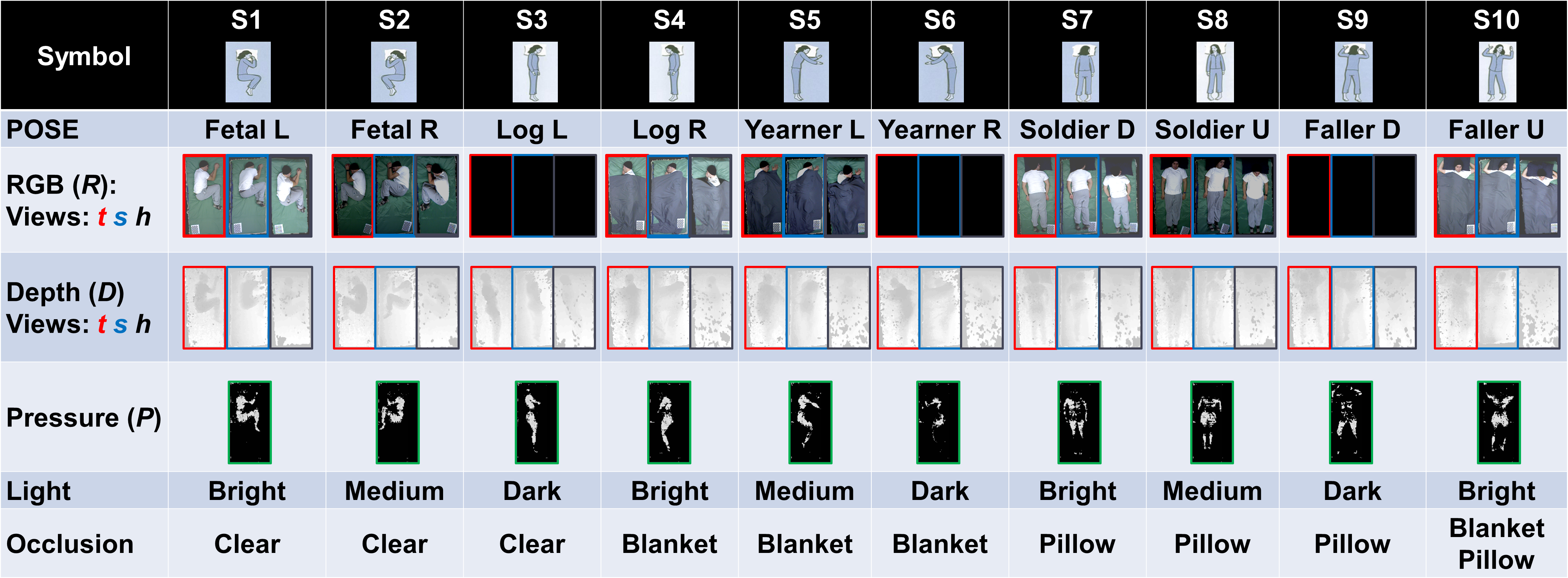}
\end{center}
   \caption{\small Multimodal and multiview dictionary of sleep poses for a single actor in various sleep configurations and scene conditions. It contains $R, D$ (equalized for display) images from $t$, $s$, and $h$ views and the pressure mat $P$. Images are transformed w.r.t the $t$ view.}
   \label{fig:MMposeDictionary}
\end{figure*}

Sample $MM$ data collected from one actor in various poses and scene conditions using all camera views and modalities is shown in Figure \ref{fig:MMposeDictionary}. The complete dataset is constructed with sleep poses collected from five actors in a mock-up ICU setting with a real ICU bed and equipment. The observations are the set of sleep poses $\mathcal{Z} =$ \{Background, Soldier U, Soldier D, Faller R, Faller L, Log R, Log L, Yearner R, Yearner L, Fetal R, Fetal L\} of size $L ~(=|\mathcal{Z}|)$ and indexed by $l$. The letters U and D indicate the patient is up-facing and down-facing and letters L and R indicate lying-on-Left and lying-on-Right sides. The variable $\mathcal{Z}_l$ is used to identify one specific pose label (e.g., $\mathcal{Z}_0$ = Background). The scene conditions are simulated using three illumination levels: bright (light sensor with 70-90\% saturation), medium (50-70\% saturation), and dark (below 50\% saturation), as well as four occlusion types: clear (no occlusion), blanket (covering 90\% of the actor's body), blanket and pillow, and pillow (between actor's head and upper back and the pressure mat). The illumination intensities are based on percent saturation values of an illumination sensor and the occlusions are detected using radio-frequency identification (RFID) and proximity sensors, all by .NET Gadgeteer. The combination of the illumination levels and occlusions types generates a 12-element scene set $\mathcal{C} =$ \{(bright, medium, dark) $\times$ (clear, blanket, pillow, blanket+pillow)\}. The variable $c \in \mathcal{C}$ is used to indicate a single illumination and occlusion combination (e.g., $c=1$ indicates bright and clear scene). The dataset is created by allowing one scene to be the combination of one actor in one pose and under a single scene condition. Ten measurements are collected from one scene -- three modalities ($R, D$, and synthetic binary masks) from each of the three camera views in the set $\mathcal{V} =\{t, h, s\}$ and one pressure image ($P$). The data collection process includes acquiring the background (empty bed), and asking the actors to rotate through the 10 poses (11 classes including the background) under each of the 12 scene conditions. The process is repeated 10 times for each of the five actors. In total, this process generates a dataset of 66,000 images (five actors $\times$ 10 sessions $\times$ 10 images $\times$ 11 classes $\times$ 12 scenes). 

\subsection{Modalities}
This section describes the modalities used by the Eye-CU system system (see Figures \ref{fig:MMposeRepresentation} and \ref{fig:MMposeDictionary}). It presents the modalities' basic properties, discusses pros and cons, and provides an intuitive justification for their complementary use in the cc-LS formulation. 

\paragraph{RGB:}
Standard RGB video data provides reliable information to represent and classify human sleep poses in scenes with relatively ideal conditions. However, most people sleep in imperfectly illuminated scenarios, using sheets, blankets, and pillows that block and disturb sensor measurements. The systems collects RGB color images of dimensions $640\times480$ from each actor in each of the scene conditions, and extracts pose appearance features representative of the lines in the human body (i.e., limbs and extremities).

\paragraph{Depth:}
Infrared depth cameras can be resilient to illumination changes. The Eye-CU system uses Primense Carmine devices to collect depth data. The devices are designed for indoor use and can acquire images of dimensions $640\times480$. These sensors use 16 bits to represent pixel intensity values, which correspond to the distance from sensor to a point in the scene. Their operating distance range is $0.8$ m to $3.5$ m; and their spatial resolution for scenes $2.0$ m away is $3.5$ mm for the horizontal (x) and vertical (y) axes, and $30$ mm along the depth (z) axis. The systems uses the depth images to represent the 3-dimensional shape of the poses. The usability of these images, however, depends on depth contrast, which is affected by the deformation properties of the mattress and blanket present in ICU environments.

\paragraph{Pressure:}
In preliminary studies, the pressure modality remained constant in the presence of sheets and blankets. The Eye-Cu systems uses the Tekscan Body Pressure Measurement System (BPMS) model BRE5315-4. The complete mat is composed of four independent pressure arrays, each with its own handle (i.e., USB adapter) to measure the pressure distribution on support surfaces. The data from each of the four arrays was synchronized and acquired using the proprietary Tekscan BPMS software. The complete pressure sensing area is $1950.7$ mm $\times 426.7$ mm with a total of 8064 sensing elements (or sensel). The sensel density is 1 sensel/cm$^2$, each with a sensing pressure range from $0$ to $250$ mm Hg (0-5 psi). The images generated using the pressure mat have dimensions of $3341 \times 8738$ pixels. Although the size of the pressure images is relatively large, the generation of such images depends on consistent physical body-mattress contact. In particular, pillows, deformation properties of the mattress, and bed configurations (not explored in this work) can disturb the measurements and the images generated by the mat. In addition, proper pressure-image generation requires a sensor array with high resolution and full bed coverage, the use of which can be prohibitively expensive and constrictive due to sanitation procedures, and limited technical support.

\subsection{Feature Extraction}
The sensors and camera views are calibrated using the standard methods from \cite{Hartley2004}. Homography transformations are computed relative to the top view and gradient and shape features are then extracted from the transformed images.

\paragraph{Histogram of Oriented Gradients (HOG).}
HOG features are extracted from RGB images to represent sleep-pose limb structures as demonstrated by \cite{dalal2005histograms, yang2013articulated}. The HOG extraction parameters are: four orientations, 16-by-16 pixels per cell, and two-by-two cells per block, which yield a 5776-element vector per image.

\begin{figure}[t]
\begin{center}
 	\includegraphics[width=.7\linewidth]{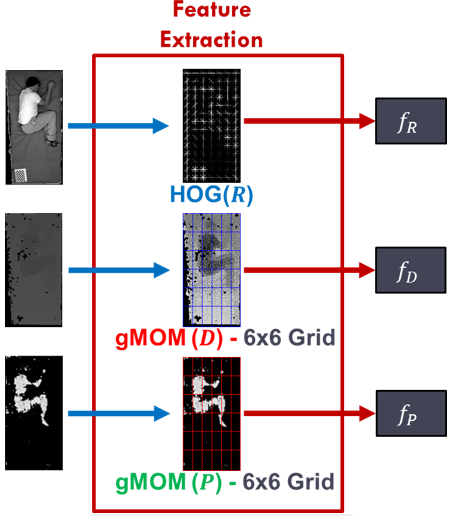}
\end{center}
  \caption{\small Multimodal representation of the Fetal L pose showing the features extracted from each modality.}
  \label{fig:FeatureExtraction}
\end{figure}

\paragraph{Geometric Moments (gMOM).}
Image gMOM features introduced in \cite{hu1962visual} and validated in \cite{ahad2012motion, ramagiri2011real} are used to represent sleep-pose shapes. The in-house implementation uses the raw pixel values from tiled depth and pressure images, instead of standard binarized pixel values. The six-by-six tile dimensions are determined empirically, to balance accuracy and complexity. Finally, moments up to the third order are extracted from each block to generate a 10-element vector per block. The vectors from each of the 36 blocks are concatenated to form a 360-element vector per image. Figure \ref{fig:FeatureExtraction} shows how features are extracted from each modality.

\section{Multimodal-Multiview Formulation}\label{sec:MMformula}
\begin{figure*}[t]
    \begin{center}
    	\includegraphics[width=.75\linewidth]{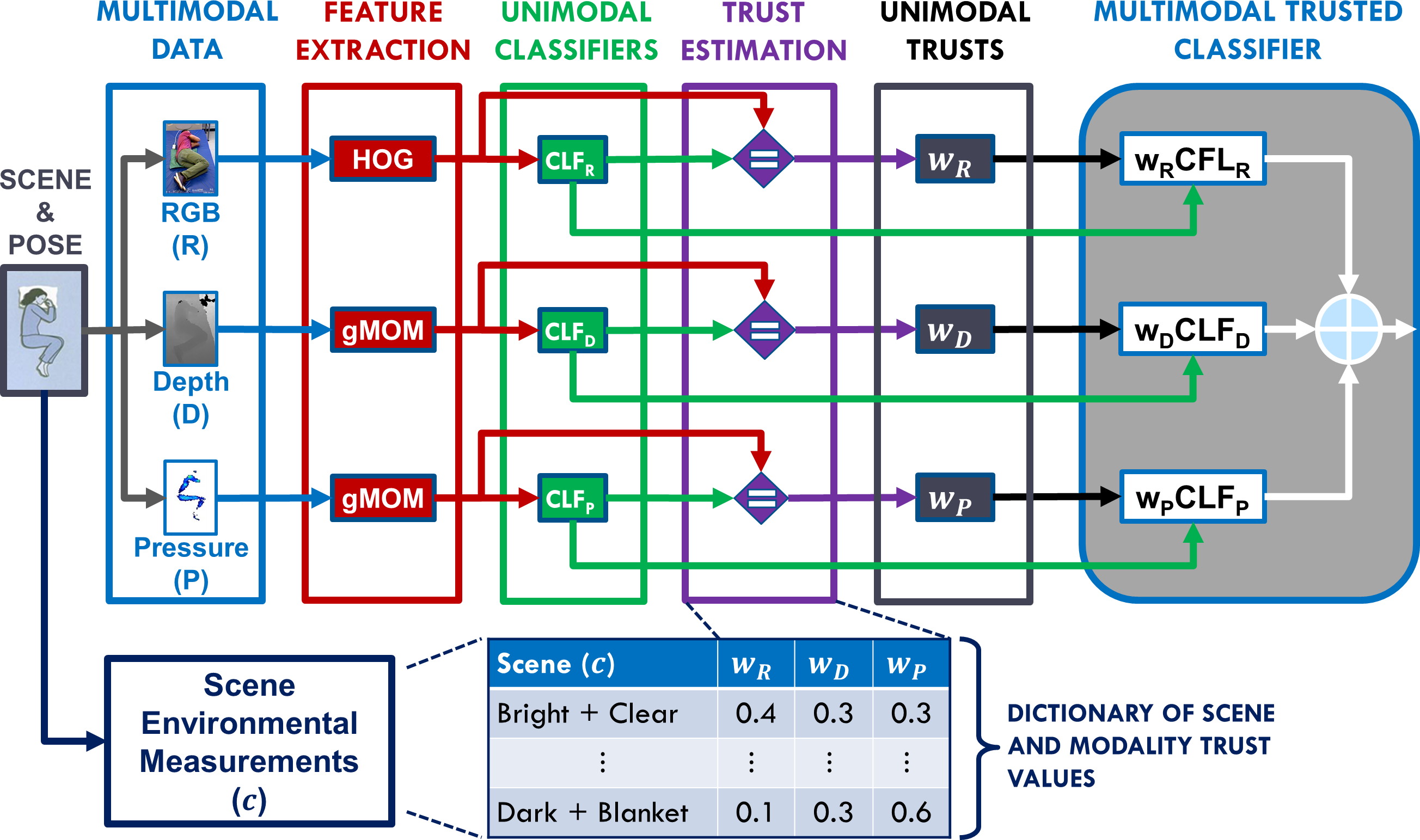}
    \end{center}
    \caption{\small Diagram of the trusted multimodal classifier for the $MpM$ configuration. Image features are extracted from the $R, D, P$ camera and pressure data. Then the features are used to train unimodal classifiers ($\text{CLF}_m$), which are in turn used to estimate the modality trust values. In the last stage of the $MM$ classifier, the unimodal decisions are trusted and combined.}
    \label{fig:MpMclassifier}
\end{figure*}

Explanation of the method begins with the problem statement in section \ref{sec:Problem}, followed by a description of the single-view multimodal formulation in section \ref{sec:construction}. This formulation is expanded to include multiview data in section \ref{sec:MultiviewFormula}. The multimodal classification framework for a single-view system is shown in Figure  \ref{fig:MpMclassifier}, which is applied to the set of pose labels $Z$ of size $L$ indexed by $l$. The multimodal dataset ($\mathcal{X}$) of size $K$ indexed by $k$ is separated for each scene $c \in \mathcal{C}$. The dataset is composed of features extracted from a set of $M$ modalities $\mathcal{N}=\{R,D,P\}$ indexed by $m$ (e.g., $f_{\mathcal{N}_m}$ with $m=1$ gives $f_R$). The $k$-th datapoint in the dataset has the form:

\begin{equation} 
    \begin{array}{ll} 
        \mathcal{X}_k & = \{f_{\mathcal{N}_m}\}_M = \{f_R, f_D, f_P\} \\
                      & = \big\{\text{HOG}(R) \text{, gMOM}(D) \text{, gMOM}(P)\big\},
    \end{array}
\end{equation}

\noindent where $f_{N_m}$ is the feature vector extracted from the $m$-th modality. These features are used to train the ensemble of $M$ unimodal SVM (and LDA) classifiers ($\text{CLF}_{m}$). For a given input datapoint $\mathcal{X}_k$, each of the classifiers outputs a probability vector $\text{CLF}_{k,m} = \begin{bmatrix} s_{k,1,m}, & ~\dots, & s_{k,L,m} \end{bmatrix}^T$, where the elements ($s$) represent the probability of label $l$ given modality feature $m$. The classifier label probabilities are computed using the implementations from \cite{scikit-learn} of Platt's method for SVC and Bayes' rule for LDA. The feature-classifier combinations are quantified at the trust estimation stage where the unimodal trust values $\mathbf{w}^c =  \begin{bmatrix} w^c_R, & w^c_D, & w^c_P \end{bmatrix}^T$ are computed for specific scene $c$. The multimodal trusted classifier is formed by fusing the candidate label decisions from the unimodal classifiers into one. The objective of this formulation is to find the pose label (${Z}_{\hat{l}}$) with the highest $MM$ probability for a given input query $\mathcal{X}_k$, where $\hat{l}$ is the estimated index label. The variables used through out this paper are listed in Table \ref{table:VariableDefinitions}. 

\subsection{Problem Statement}\label{sec:Problem}
The proposed fusion technique uses probabilistic concepts to compute the probability of a given class by marginalizing the joint probability over the modalities. The joint probability is calculated from the conditional probability of each class and the set of prior probabilities for each modality. The conditional probabilities are extracted from the classifiers in the ensemble of $M$-unimodal classifiers (i.e., $\mathbb{P}(Z = Z_l | \mathcal{X} = \mathcal{X}_k) = \mathbb{P}(Z_l|\mathcal{X}_k)$) and re-written as:
\begin{equation}
    \mathbb{P}(Z_l|\mathcal{X}_k) = \sum_{m=1}^M \mathbb{P}(Z_l | \mathcal{X}_k, M = m) \mathbb{P}(M = m).
\end{equation}

Methods such as Platt's ~\cite{platt1999fast} for SVMs enable the computation of conditional probabilities given by: 
\begin{equation}
    s_{k, l, m} = \mathbb{P}(Z_l | \mathcal{X}_k, M = m).
\end{equation} 

\noindent However, the prior probability for each modality $w_m = \mathbb{P}(M = m)$ remains unknown. The trust method finds the set of priors for each modality $m$ in the ensemble of $M$ modalities that approximates the following probability:

\begin{equation}
    b_{ k,l} = \mathbb{P}(Z = z_l | \mathcal{X} = \mathcal{X}_k, \text{Oracle}),
\end{equation}

\noindent produced by an oracle-observed  datapoint $\mathcal{X} = \mathcal{X}_k$.
The estimation process is repeated for all $c$'s. However, $c$ is omitted to simplify the notation (i.e., $\mathbf{w}^c$ becomes $\mathbf{w}$).

The method uses the following coupled optimization problem to find the modality priors $w_m$ for scene $c$: 
\begin{equation}
    \begin{aligned}
        & \underset{\mathbf{w}}{\text{minimize}}
        & & \frac{1}{2}  \sum_{k=1}^K   \sum_{l=1}^L  \left(\sum_{m=1}^M s_{k, l, m} w_m - b_{k, l}\right)^2\\
        & \text{subject to}
        & & \mathbf{1}^T \mathbf{w} =1 \\
        & & & 0 \leq w_m \leq 1, m=1,\hdots,M,
    \end{aligned}
    \label{eq:general_optimization}
\end{equation}
The objective is to find the weights $w_m$ that approximate the oracle $b_{k, l}$ for every data point $\mathcal{X}_k$. Using the loss in Eq.~\ref{eq:general_optimization}, the problem becomes a cc-LS optimization problem. This type of problem uses all points and poses labels from the training set to find the set of priors that approximates the values produced by the oracle for each point $\mathcal{X}_k$ at once.

\begin{table}
\small
\centering
\begin{tabular}{l||l}
\hline
\multicolumn{2}{c}{\bf{VARIABLES}}               \\ \hline\hline
SYMBOL            & DESCRIPTION                                                 \\ \hline
$\mathbf{A}$      & Multimodal matrix $\in \mathbb{R} ^{U \times M}$ \\
$\mathbf{a}_m$    & $m$-th column vector of $\mathbf{A}$ with $U$ elements      \\
$\mathbf{b}$      & Oracle vector $\in \mathbb{R}^ U$                           \\
$\mathbf{b}_m$    & Oracle column vector for modality $m$\\
$\mathcal{C}$     & Scenes set (light $\times$ occlusion) combination           \\
$c$               & Scene index, $1\leq c \leq |\mathcal{C}|$                   \\
$\text{CLF}_{k,m}$& Classifier for the $m$-th modality                          \\
$\{f_{\mathcal{N}_m}\}_k$   & Set of feature $M$ vectors for the $k$-th datapoint       \\
$D$               & Depth modality                                              \\
$h$               & Head camera view                                            \\
$K$               & Dataset size, $K = |\mathcal{X}|$                           \\ 
$k$               & Datapoint index, $1\leq k \leq K$                           \\
$L$               & Size of the set of pose labels, $L = |\mathcal{Z}|$         \\
$l$               & Index of the pose label ($\mathcal{Z}_l$), $1\leq l \leq L$ \\
$\hat{l}$         & Index of the estimated pose label, $1\leq \hat{l} \leq L$   \\
$l^*$             & Index of the ground truth label ($Z_l$), $1\leq l^* \leq L$ \\
$MM$              & Multimodal and Multiview                                    \\
$MpM$             & Multimodal and partial-Multiview                            \\
$M$               & Size of modality set, $M = |N|$                             \\
$m$               & Modality index, $1\leq m \leq M$                            \\
$\mathcal{N}$     & Modality set $\mathcal{N} = \{R, D, P\}$ indexed by $m$     \\
$P$               & Pressure modality                                           \\
$pMM$             & Partial-Multimodal and Multiview                            \\
$PMpM$            & Partial-Multimodal partial-Multiview                        \\
$R$               & RGB modality                                                \\
$s$               & Side camera view                                            \\ 
$s_{k,l,m}$       & Probability of label $l$ from $\text{CLF}_{k,m}$            \\
$t$               & Top camera view                                             \\
$U$               & Multimodal dimension $U = KL$                               \\
$\mathcal{V}$     & View set $\mathcal{V}=\{t, s, h\}$ \\
$V$               & Number of views $V= |\mathcal{V}|$                          \\
$v$               & View index, $1\leq v \leq V$                                \\
$\mathbf{w}^{c}$  & Trusts $\mathbf{w}=\begin{bmatrix} w_R, w_D, w_P \end{bmatrix}^T$ for scene $c$ \\
$w_{N_m}$         & Modality trust value (e.g., $w_{R}$ for $m=1$)              \\ 
$\mathcal{X}$     & Dataset indexed by $k$ (i.e., $\mathcal{X}_k$)              \\
$\mathcal{X}_k$   & $k$-th datapoint with $\{f_{N_m}\}_k = \{f_R, f_D, f_P\}_k$ \\
$Y$               & $MM$ dimensions ($=K L V$)        \\
$\mathcal{Z}$     & Sleep-pose set \\
\hline
\end{tabular}
\caption{Variables and their descriptions.}
\label{table:VariableDefinitions}
\end{table}

\subsection{Multimodal Construction}\label{sec:construction}
The estimation method uses cc-LS optimization to minimize the difference between Oracle ($\mathbf{b}$) and the multimodal matrix ($\mathbf{A}$). It frames the trust estimation as a linear system of equations of the form $\mathbf{Aw} - \mathbf{b} = 0$, where the modality trust values are the elements of the vector $\mathbf{w}=\begin{bmatrix}w_R, & w_D, & w_P\end{bmatrix}^T$ that approximate $\mathbf{A w}$ to $\mathbf{b}$. 

\paragraph{Construction of the Multimodal Matrix ($\mathbf{A}$).}\label{sec:Aconstruction}
The matrix $\mathbf{A}$ contains label probabilities for each of the datapoints in the training set ($K=|\mathcal{X}_{train}|$). This matrix has $U$ rows ($U = K L$) and $M$ columns, where $L$ is the total number of labels ($L=|\mathcal{Z}|$), and $M$ is the number of modalities ($M=3$) and has the following structure:
\begin{equation}\label{eqn:A_matrix}
    \begin{aligned}
      \mathbf{A} & = \begin{bmatrix}
                       \mathbf{S}_{k=1}^T, & \hdots, & \mathbf{S}_{k=K}^T,
                       
                     \end{bmatrix}^T_{U \times M}
    \end{aligned}
\end{equation}

\noindent where $\mathbf{S}_{k}(l,m) = s_{k, l, m}$.

\paragraph{Construction of the Multimodal Oracle Vector ($\mathbf{b}$).}\label{sec:bconstruction}
The vector $\mathbf{b}$ is generated by the oracle and quantifies the classification ability of the combined modalities. It is used to corroborate estimation correctness when compared to the ground truth. The $\mathbf{b}_m$ column vectors have $U$ rows: 
\begin{equation}\label{eqn:b_vector}
    \mathbf{b}_m = 
        \begin{bmatrix}
            \mathbf{b}_{k=1}^T, & \hdots, & \mathbf{b}_{k=K}^T
        \end{bmatrix}^T,
\end{equation}

\noindent where $\mathbf{b}_k  = \begin{bmatrix}b_{k,l=1}, & \hdots, & b_{k, l=L} \end{bmatrix}^T$. The values of the $b_{k,l}$ elements are set using the following condition:

\begin{equation}\label{eqn:b_if}
    b_{k,l} =\Bigg\{\begin{array}{ll} 
                    1, & \text{if } \hat{l} = l^* \text{ for } \mathcal{X}_k\\
                    0, & \text{otherwise, } \\
                   \end{array}
\end{equation}

\noindent where $\hat{l} = {\arg\max} ~s_{k,l,m}$ is the index of the estimated label and $l^*$ is the index of the ground truth label for $\mathcal{X}_k$. 

The construction of the oracle $\mathbf{b}$ depends on how the columns $\mathbf{b}_m$ (i.e., unimodal oracles) are combined. The system is tested with a uniform construction and the results are reported in section \ref{sec:Experiments}. In the uniform construction, each modality has a $\frac{1}{M}$ voting power and can add up to one via:

\begin{equation}\label{eqn:b_vector_uniform}
    \mathbf{b} = \frac{ {\underset{\forall m}{\sum}} \mathbf{b}_m}{M}. 
\end{equation}

\subsection{Coupled Constrained Least-Squares (cc-LS)}
Finally, the weight vector $\mathbf{w} = [w_R, w_D, w_P]^T$ is computed by substituting $\mathbf{A}$ and $\mathbf{b}$ into Eqn. (\ref{eq:general_optimization}) and solving the cc-LS optimization problem: 
\begin{equation}\label{eq:boundedLS}
    \begin{aligned}
    & \underset{\mathbf{w}}{\text{minimize}}
    & & \frac{1}{2} \|\mathbf{A} \mathbf{w} - \mathbf{b} \|_2^2  \\
    & \text{subject to}
    & & \mathbf{1}^T \mathbf{w} =1 \\
    & & & 0 \leq w_m \leq 1, m=1,\hdots,M
    \end{aligned}.
\end{equation}

Intuitively, the cc-LS problem finds the modality priors that allow the method to fuse information from different modalities to approximate the oracle probabilities.

\subsection{Multiview Formulation}\label{sec:MultiviewFormula}
The bounded multimodal formulation is expanded to include multiview data using $V$ views indexed by $v$. The values that $v$ can take indicate which camera view is used (e.g., $v=1$ for the top view, $v=2$ for the side view, and $v=3$ for the head view.) The multimodal and multiview matrix $\mathbf{A}$ has the following form:

\begin{equation}\label{eqn:A_matrix_view}
\mathbf{A} = \begin{bmatrix}[\mathbf{A}^{(v=1)}], \dots, [\mathbf{A}^{(v=V)}]\end{bmatrix}_{Y\times M}^T,
\end{equation}

where $Y = L K V$ for a system with $V$ views and $M$ modalities. The $\mathbf{b}_m$ multimodal and multiview oracle vector is constructed by concatenating data from all the views in the set ($\mathcal{V}$) via:

\begin{equation}
\mathbf{b}_m = 
        \begin{bmatrix}
            \begin{bmatrix}
                \mathbf{b}_{k=1}^{(v=1)}
            \end{bmatrix}^{T}, 
            \hdots,  
            \begin{bmatrix}
                \mathbf{b}_{k=1}^{(v=V)}
            \end{bmatrix}^{T}
        \end{bmatrix}_{Y}^T,
\end{equation}

\noindent and the $\mathbf{b}$ column vector is generated using (\ref{eqn:b_vector_uniform}).

\subsection{Testing}
\begin{figure*}[t]
    \begin{center}
    	\includegraphics[width=.85\linewidth]{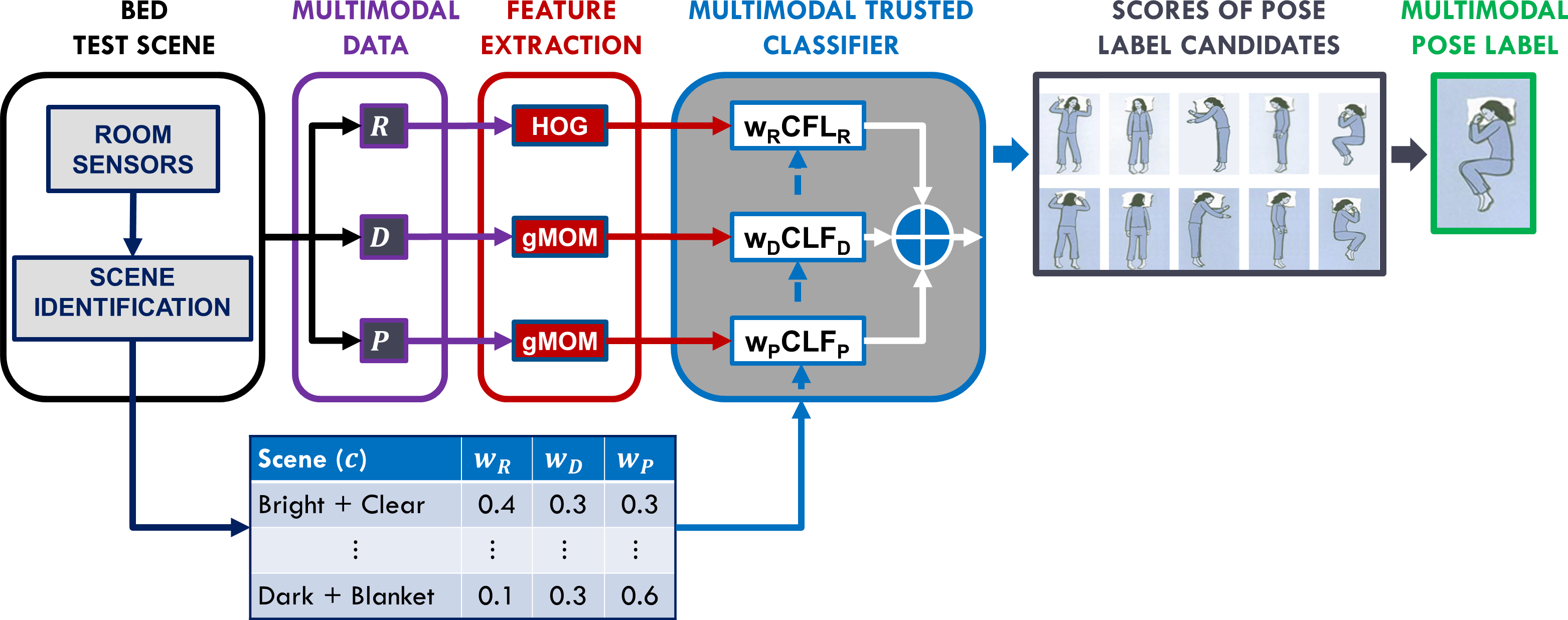}
    \end{center}
    \caption{\small Block diagram for testing of a single view multimodal trusted classifier. Observations ($R,D,P$) are collected from the scene. Features are extracted from the observations and sent to the unimodal classifiers to provide a set of score-ranked pose candidate labels. The set of candidates is trusted and combined into one multimodal set from which one with the highest score is selected.}
    \label{fig:testing}
\end{figure*}

The test process is shown in Figure  \ref{fig:testing}. The room sensors in combination with $ \mathcal{N}=\{R,D,P\}$ measurements are collected from the ICU scene. Features ($\{f_{N_m}\}_k$) are extracted from the modalities in $\mathcal{N}$ and are used as inputs to the trusted multimodal classifier. The classifier outputs a set of label candidates from which the label with the largest probability for datapoint $\mathcal{X}_k = \{f_{N_m}\}_k$ is selected via: 

\begin{equation}
    {\hat{l}_k} = \underset{l \in L} {\arg\max} \left( w_{N_{m}} \text{CLF}_m \{f_{N_m}\}_k \right), \forall m.
\end{equation}

\paragraph{Missing Modalities}
Hardware failures are simulated by evaluating the classification performance with one modality removed at a time. The trust value of a missing or failing sensor modality ($w^*_{N_n}$) is set to zero and its original value ($w_{N_n}$) is proportionally distributed to the others via:

\begin{equation}\label{eq:modeAdjust}
        w^*_{N_m} = w_{N_m}  \left(1 + \frac{|w_{N_n} - w_{N_m}|}{W}\right), 
\end{equation}
\noindent for $n \in \{1, ..., M\}$, $m \in \{1, ..., M\}\setminus n$, and $W = \underset{\forall m}{\sum} w_m.$

\section{Experiments}\label{sec:Experiments}
Validation of modalities and views for sleep-pose classification substantiates the need for a multiview and multimodal system. The cc-LS method is tested on the $MpM$, $MM$, $PMM$ and $PMpM$ Eye-CU configurations and data collected from scenes with various illumination levels and occlusion types. The labels are estimated using multi-class linear SVC (C=0.5) and LDA classifiers from \cite{scikit-learn}. A validation set is used to tune the SVC's C parameter and the Ada parameters. Classification accuracies are computed using five-fold cross validation using in-house implementations of competing methods and reported as percent accuracy values inside color-scaled cells.

\subsection{Modality and View Assessment}
\begin{figure*}
    \begin{center}
    	\includegraphics[width=.85\linewidth]{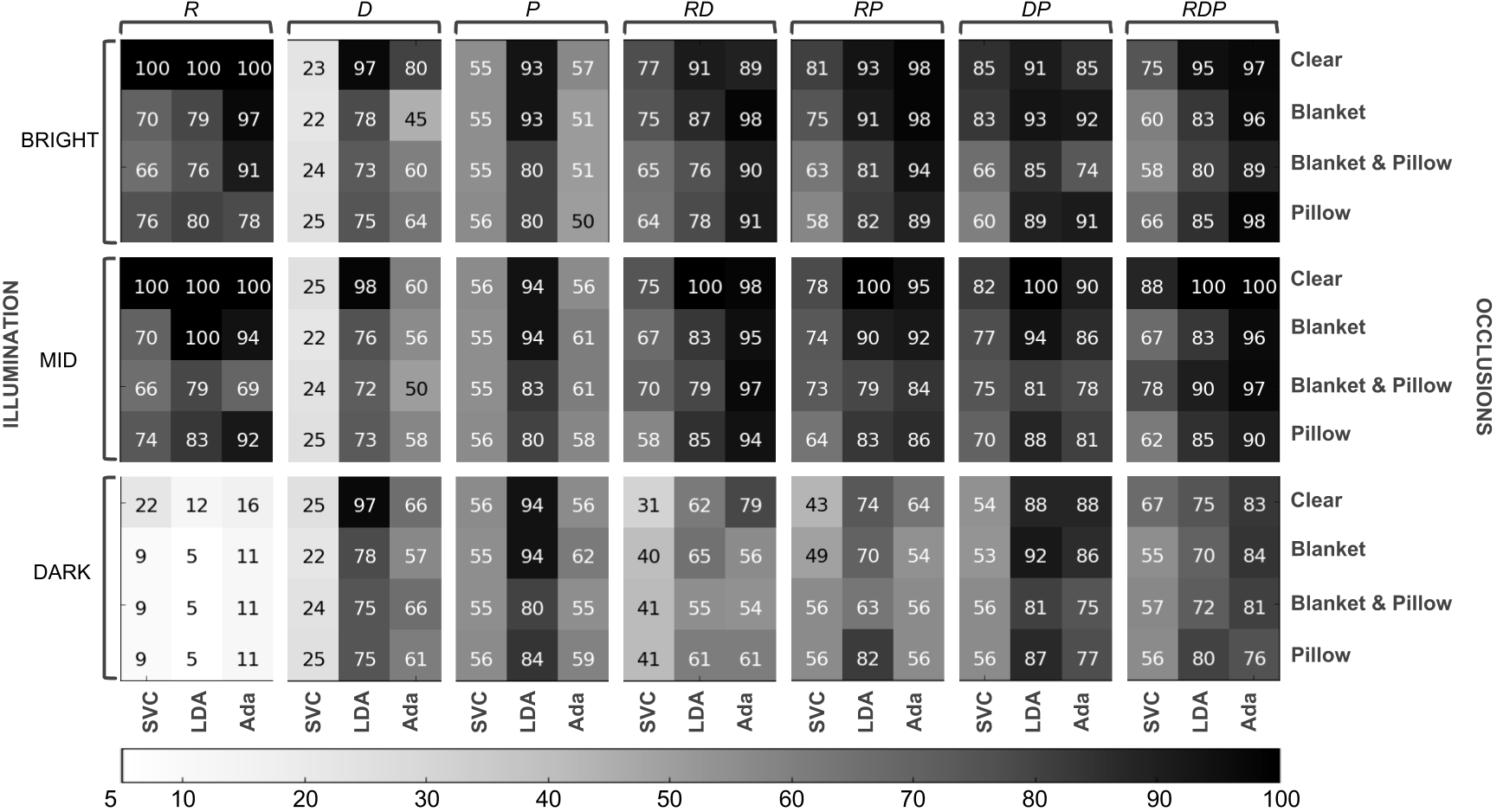}
    \end{center}
    \caption{ \small Performance evaluation of modalities and modality combinations using SVC, LDA, and Ada-Boosted SVC (Ada) based on their classification percent accuracy (cell values). The evaluation is performed over all the scene conditions considered in this study. The results indicate that no single modality ($R, D, P$) or combination of concatenated modalities ($RD, RP, DP, RDP$) in combination with one of three classification techniques cannot be directly used to recognize poses in all scenes. The top row indicates which modality or combination of modalities is used. The labels on the bottom indicate which classifier is used. The labels to the left and right indicate the scene's illumination level and occlusion types. The gray-scaled boxes range from worst (white) to best (black) performance.}
    \label{fig:ModalityEval}
\end{figure*}

Classification results obtained using unimodal and multimodal data without modality trust are shown in Figure \ref{fig:ModalityEval}. The cell values indicate classification percent accuracy for each individual modality and modality combinations with three common classification methods. The labels of the column blocks at the top of the figure indicates modalities used. The labels at the bottom of the figure show which classifier is used. The labels on the left and right indicate scene illumination level and type of occlusion. The figure only shows classification results for the top camera view because variation across views tested did not have statistical significance.

\subsection{Performance of Reduced Eye-CUs}\label{sec:exp_configurations}
\begin{figure}
    \begin{center}
    	\includegraphics[width=1\linewidth]{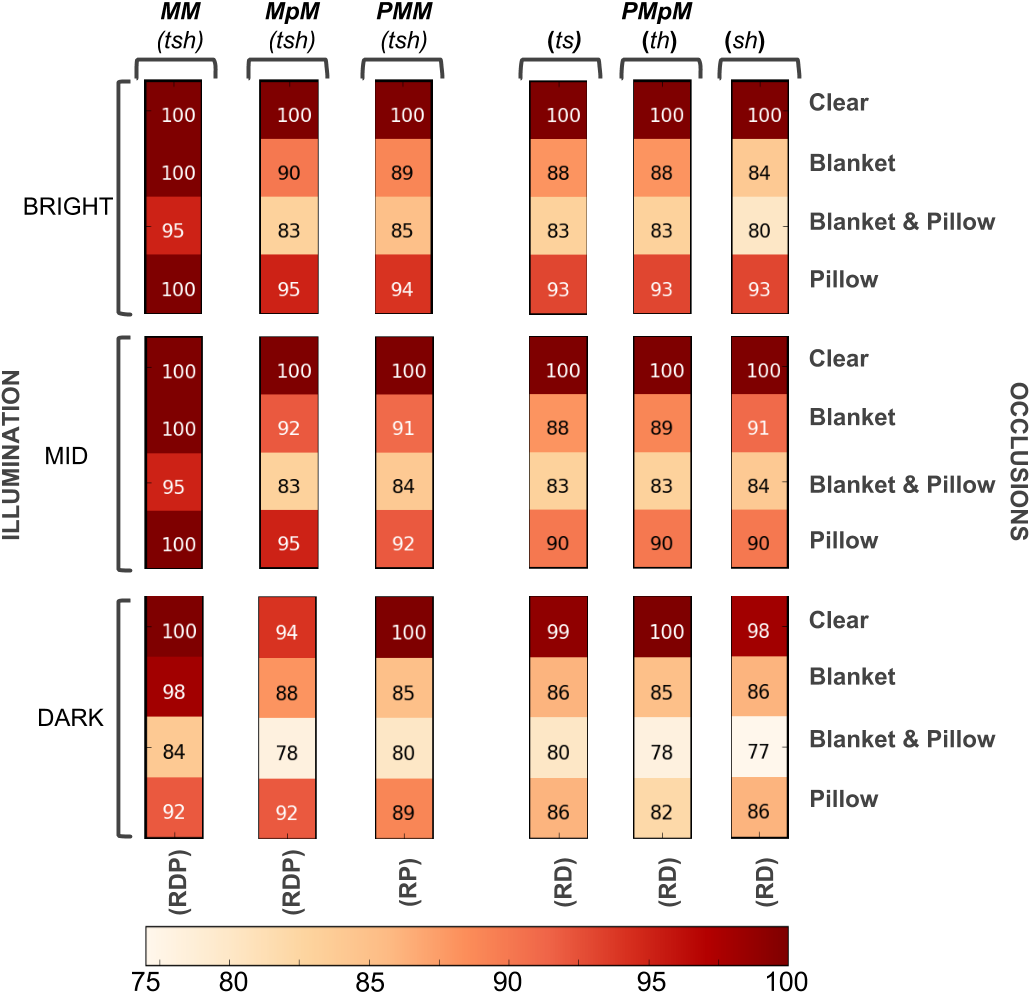}
    \end{center}
    \caption{\small Classification performance in red scale (dark: best, light: worst) of the various Eye-CU configurations using LDA. The $PMpM$ has the lowest performance of 76.7\% using $sh$ views of a dark and occluded scene. The method from \cite{torres2015Multimodal} performs below 50\% and the method from \cite{huang2010multimodal} is not suited for such conditions. The top row identifies the configuration. The second row indicates views used. The bottom labels indicate modalities used (in parenthesis). The labels on the left and right indicate scene illumination and occlusion type. Similar pattern is observed with SVC.}
    \label{fig:SysConfigurations}
\end{figure}

The complete $MM$ configuration achieves the best classification performance, followed closely by the performances of the $MpM$, $PMM$, and $PMpM$ configurations, which is summarized in figure  \ref{fig:SysConfigurations}. The values inside the cells represent classification percent accuracy of the  cc-Ls method combined with various Eye-CU system configurations. The top row indicates the configuration. The second row indicates the views. The labels on the bottom of the figure identify the modalities. The labels on the left and right indicate illumination level and occlusion type. The red scale ranges from light red (worst) to dark red (best). The figure shows that the complete $MM$ system in combination with the cc-LS method performs the best across all scenes. However, it requires information from a pressure mat. The $PMM$ and $PMpM$ configurations do not require the pressure mat and are still capable of performing reliably and with only a slight drop in their performance. For example, in dark and occluded scenes the $PMM$ and $PMpM$ configurations reach 77\% and 80\% classification rates respectively (see row: DARK; Blanket \& Pillow). 

\subsection{Comparison with Existing Methods}
\begin{figure}
    \begin{center}
    	\includegraphics[width=1\linewidth]{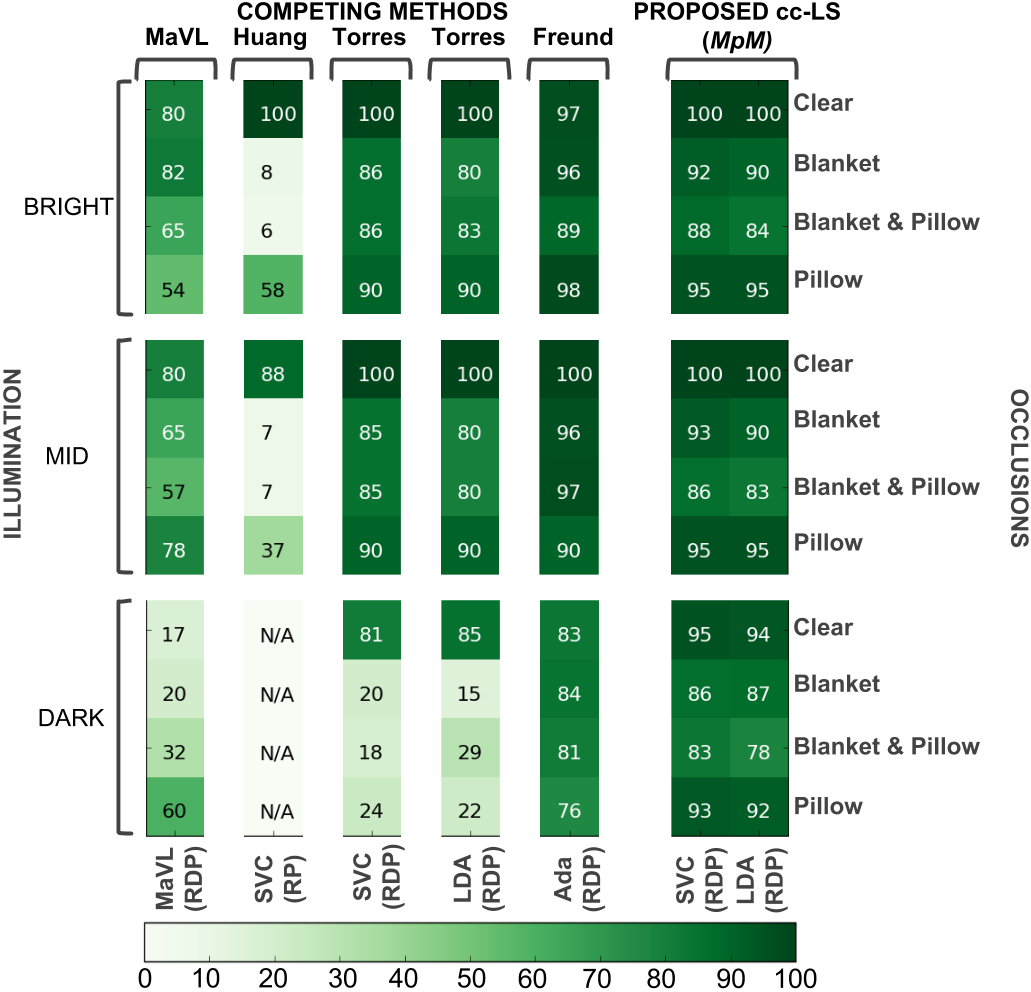}
    \end{center}
    \caption{\small Mean classification performance in green scale (dark: best, light:worst) of MaVL, Huang's \cite{huang2010multimodal}, Torres' \cite{torres2015Multimodal}, Feund's \cite{freund1997decision} and the cc-LS method using SVC and LDA. The combination of cc-LS and $MpM$ matches the performance of competing methods in bright and clear scenes. Classification is improved with cc-LS by 70\% with SVC and by 30\% with LDA in dark and occluded scenes. The top row distinguishes between competing and proposed methods; the second row cites them. The bottom row indicates classifier and modalities (in parenthesis) used. The labels on the left and right indicate scene illumination and occlusion type. N/A indicates not suitable.}
    \label{fig:ComparedMethods}
\end{figure}

Performance of the cc-LS and the in-house implementations of the competing methods from \cite{huang2010multimodal}  and \cite{torres2015Multimodal} and Ada \cite{freund1997decision} are shown in Figure  \ref{fig:ComparedMethods}. The figure shows results using the $MpM$ configuration, which more closely resembles those used in the competing methods. All the methods use a multimodal system with a top camera view and a pressure mat. The values inside the cells are the classification percent accuracy. The green scale goes from light green (worst) to dark green (best). The top row divides the methods into competing and proposed. The second row cites the methods. The bottom row indicates which classifier and, in parentheses, modalities are used. The labels on the left and right indicate illumination level and occlusion type. The results are obtained using the four methods with $MM$ dataset.

\begin{figure}
   \begin{subfigure}{1\linewidth}
      \includegraphics[width=1\linewidth]{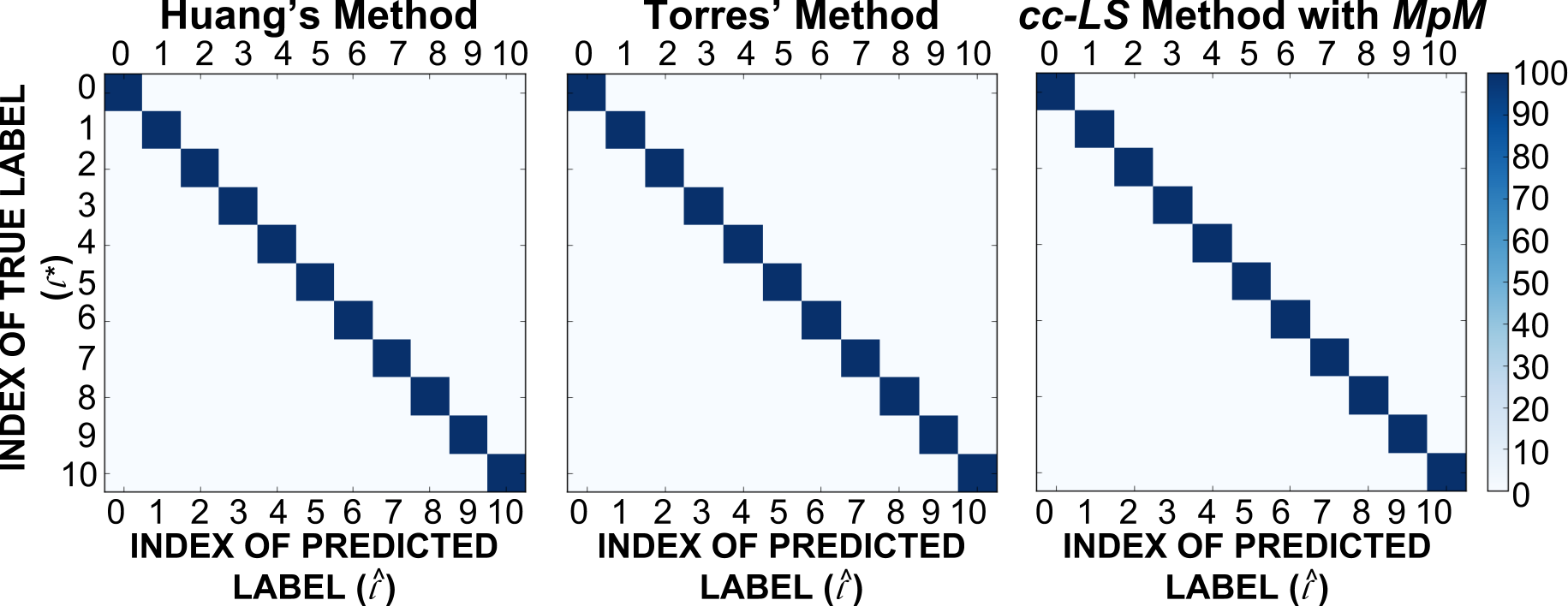} 
      \caption{\small Bright scene clear of occlusions.}
      \label{fig:cmIdeal}
   \end{subfigure}
   \begin{subfigure}{1\linewidth}
      \includegraphics[width=1\linewidth]{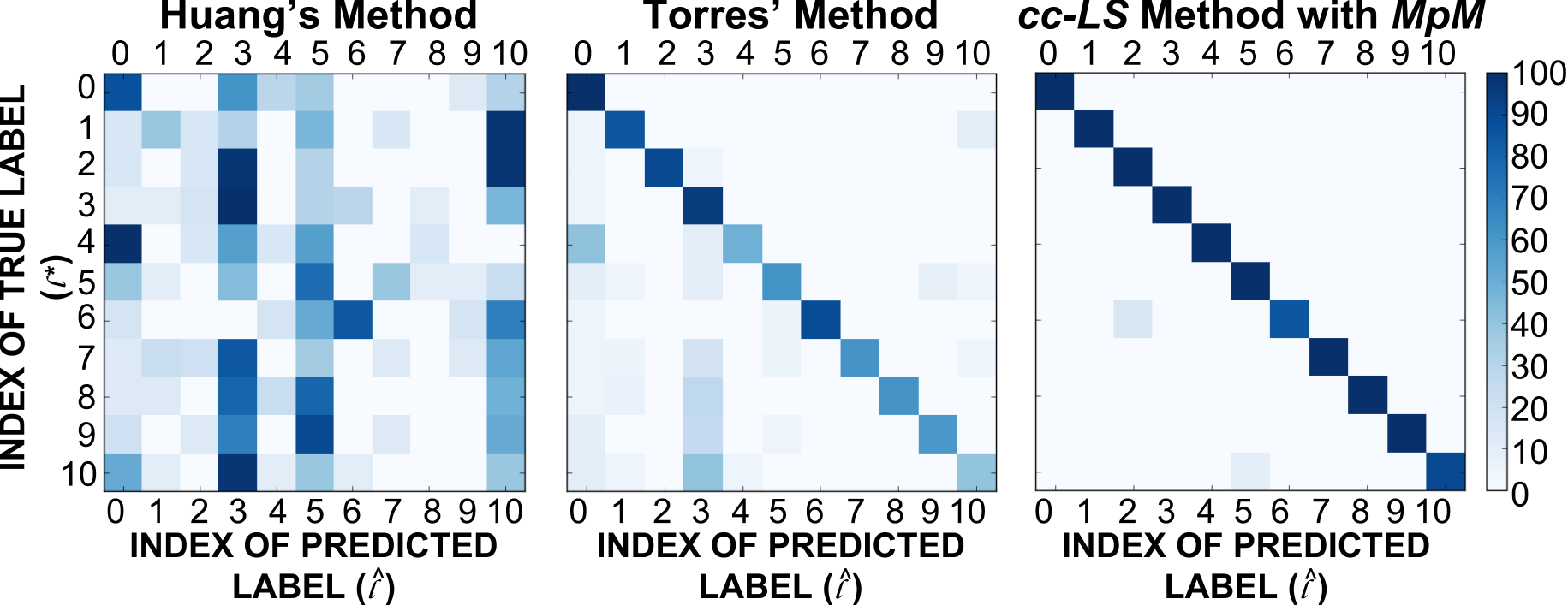}
      \caption{\small Dark scene with pillow and blanket occlusions.}
      \label{fig:cmPillowBlanket}
   \end{subfigure}
\caption{\small Confusion matrices generated in blue scale (dark: best, light: worst) using a top camera view and applying the methods from Huang's \cite{huang2010multimodal}, Torres' \cite{torres2015Multimodal}, and cc-LS with $MpM$. The top matrices show all methods have perfect classification in ideal scenes (i.e., main diagonal). The bottom matrices are \cite{huang2010multimodal} with 7\%, \cite{torres2015Multimodal} with 55\%, and cc-LS with 86.7\% for dark and occluded scenes. The matrices show the matches between estimated ($\hat{l}$) and ground truth ($l^*$) indices.}\label{fig:CMs}
\end{figure}

\paragraph{Confusion Matrices:} The confusion matrices in Figure  \ref{fig:CMs} show how the indexes of estimated labels $\hat{l}$ match the actual labels $l^*$. The top three matrices are from a scene with bright and clear ICU conditions (Figure \ref{fig:cmIdeal}). The bottom three matrices illustrate the performance of the methods in a dim and occluded ICU scenario (Figure \ref{fig:cmPillowBlanket}). A dark blue diagonal in the confusion matrices indicates perfect classification. In the selected scenes, all methods achieved a 100 \% classification for the bright and clear scene. However, their performance varies greatly in dim and occluded scenes. The matrix generated using \cite{huang2010multimodal} achieves 7\% classification accuracy (bottom left), matrix generated using \cite{torres2015Multimodal} achieves a 55\% accuracy (bottom center), and the matrix generated with the cc-LS method achieves a 86.7\% accuracy (bottom right). The $MpM$ configuration with the cc-LS method outperforms the competing methods by an approximate 30\%.

\paragraph{Performance of Ada-Boost}
The system is tested using Ada-Boost (Ada) algorithm \cite{freund1997decision} to improve the decision of weak unimodal SVCs. The results from Figure  \ref{fig:ModalityEval} show a slight SVC improvement. The  comparison in Figure \ref{fig:ComparedMethods} shows that the Ada's improvement is small. It barely outperforms the reduced $MpM$ configuration with cc-LS method in some scenes (see row: MID-Blanket). Overall, Ada is outperformed by the combination of cc-LS and $MpM$.

\section{Discussion}\label{sec:Discussion}
The results in Figure  \ref{fig:ComparedMethods} show performance disparities between the results obtained with the in-house implementation and those reported in \cite{huang2010multimodal}. The data and code from \cite{huang2010multimodal} were not released, so the findings and implementation details reported in this paper cannot be compared at the finest level. Nevertheless, the accuracy variations observed are most likely due to differences in data resolutions, sensor capacities, scene properties, and tuning parameters.

The performance of the $MM$ and $MpM$ configurations, which use a pressure mat, is slightly improved. However, the deployment and maintenance of such systems in the real world can be very difficult and perhaps logistically impossible. The cc-LS method in combination with the $PMM$ or $PMpM$ configurations, which do not use a pressure mat, match and outperform the competing techniques in ideal and challenging scenarios (see Figure  \ref{fig:ComparedMethods}).

\section{Conclusion and Future Work}\label{sec:Conclusion}
This work introduced a new modality trust estimation method based on cc-LS optimization. The trust values approximate the difference between the multimodal candidate labels $\mathbf{A}$ and the expected oracle $\mathbf{b}$ labels. The Eye-CU system uses the trust to weight label propositions of available modalities and views. The cc-LS method with the $MM$ Eye-CU system outperforms three competing methods. Two reduced Eye-CU variations reliably classify sleep poses without pressure data. The $MM$ properties allow the system to handle occlusions and avoid problems associated with a pressure mat (e.g., sanitation and sensor integrity).

Reliable pose classification methods and systems enable clinical researchers to design, enhance, and evaluate pose-related healthcare protocols and therapies. Given that the Eye-CU system is capable of reliably classifying human sleep poses in an ICU environment, expansion of the system and methods is under investigation to include temporal information. Future analysis will seek to quantify and typify pose sequences (i.e., duration and transition). Future work will investigate removing the constraints that clearly define the set of sleep poses and explore tools from novelty detection to identify other (e.g., helpful and harmful) patient poses that occur in an ICU. Recent studies indicate that deep features might improve the classification performance of the Eye-CU system in the most challenging healthcare scenarios. Hence, future work will investigate the performance and integration of deep features into the cc-LS method and the Eye-CU system.

\paragraph*{Acknowledgements}
 This project was supported in part by the Institute for Collaborative Biotechnologies (ICB) through grant W911NF-09-0001 from the U.S. Army Research Office; and by the U.S. Office of Naval Research (ONR) through grant N00014-12-1-0503. The content of the information does not necessarily reflect the position or the policy of the Government and no official endorsement should be inferred.
 
\newpage
\footnotesize
\bibliographystyle{ieee}

\end{document}